\documentclass[10pt,twocolumn,letterpaper]{article}

\usepackage{cvpr}
\usepackage{times}
\usepackage{epsfig}
\usepackage{graphicx}
\usepackage{amsmath}
\usepackage{amssymb}
\usepackage{algorithm}
\usepackage{multirow}
\usepackage{mathtools,xparse}
\usepackage{subcaption}
\usepackage{siunitx}
\usepackage{stmaryrd}
\usepackage[noend]{algpseudocode}
\usepackage{placeins}
\usepackage{mdframed}

\usepackage{amsthm}

\algnewcommand\algorithmicforeach{\textbf{for each}}
\algdef{S}[FOR]{ForEach}[1]{\algorithmicforeach\ #1\ \algorithmicdo}

\usepackage[pagebackref=true,breaklinks=true,letterpaper=true,colorlinks,bookmarks=false]{hyperref}

\newcommand*\rot{\rotatebox{90}}
\newcommand{\emptycol}{\multicolumn{1}{c|}{--}}
\cvprfinalcopy 

\def\cvprPaperID{362} 

\begin{document}

\title{Graph-Cut RANSAC}

\author{Daniel Barath\\
Machine Perception Research Laboratory\\
MTA SZTAKI, Budapest, Hungary\\
{\tt\small barath.daniel@sztaki.mta.hu}
\and
Jiri Matas\\
Centre for Machine Perception, Department of Cybernetics \\
Czech Technical University, 
Prague, Czech Republic\\
{\tt\small matas@cmp.felk.cvut.cz}
}

\maketitle

\definecolor{schematic_figure_border_color}{rgb}{.7,.7,.7}
\mdfdefinestyle{schematic_figure}{%
	linewidth = 0.8pt,%
	linecolor = schematic_figure_border_color,%
    shadowsize = 0pt,
}

\begin{abstract}
	A novel method for robust estimation, called Graph-Cut RANSAC\footnote{Available at \url{https://github.com/danini/graph-cut-ransac}}, GC-RANSAC in short, is introduced. To separate inliers and outliers, it runs the graph-cut algorithm in the local optimization (LO) step which is applied when a \textit{so-far-the-best} model is found. The proposed LO step is conceptually simple, easy to implement, globally optimal and efficient. 
GC-RANSAC is shown experimentally, both on synthesized tests and real image pairs, to be more geometrically accurate than state-of-the-art methods on a range of problems, e.g.\ line fitting, homography, affine transformation, fundamental and essential matrix estimation. It runs in real-time for many problems at a speed approximately equal to that of the less accurate alternatives (in milliseconds on standard CPU).  

%
\end{abstract}

\section{Introduction}

The RANSAC (RANdom SAmple Consensus) algorithm proposed by Fischler and Bolles~\cite{fischler1981random} in 1981 has become the most widely used robust estimator in computer vision. RANSAC and similar \textit{hypothesize-and-verify} approaches have been successfully applied to many vision tasks, e.g.\ to short baseline stereo~\cite{torr1993outlier,torr1998robust}, wide baseline stereo matching~\cite{pritchett1998wide,matas2004robust,mishkin2015mods}, motion segmentation~\cite{torr1993outlier}, image mosaicing~\cite{ghosh2016survey}, detection of geometric primitives~\cite{sminchisescu2005incremental}, multi-model fitting~\cite{zuliani2005multiransac}, or for initialization of multi-model fitting algorithms~\cite{isack2012energy,pham2014interacting}.
In brief, the RANSAC approach repeatedly selects random subsets of the input data and fits a model, e.g.\ a line to two points or a fundamental matrix to seven point correspondences. In the second step, the model support, i.e.\ the number of inliers, is obtained. The model with the highest support, polished e.g.\ by a least squares fit on inliers, is returned.

In the last three decades, many modifications of RANSAC have been proposed. For instance, NAPSAC~\cite{nasuto2002napsac}, PROSAC~\cite{chum2005matching} or EVSAC~\cite{fragoso2013evsac} modify the sampling strategy to increase the probability of selecting an all-inlier sample earlier. 
NAPSAC considers spatial coherence in the sampling of input data points, PROSAC exploits the ordering of the points by their predicted inlier probability, EVSAC uses an estimate of the confidence in each point. Modifications of the model support step has also been proposed.
In  MLESAC~\cite{torr2000mlesac} and MSAC~\cite{hartley2003multiple},  the model quality is estimated by a maximum likelihood process, albeit under certain assumptions, with all its beneficial properties. In practice, MLESAC results are often superior to the inlier counting of plain RANSAC and less sensitive to the user-defined threshold. The termination of RANSAC is controlled by a manually set confidence value $\eta$ and the sampling stops when the probability of finding a model with higher support falls below $\eta$\footnote{This interpretation of $\eta$ holds for the standard cost function only.}.   

Observing that RANSAC requires in practice  more samples than theory predicts, Chum et al.~\cite{chum2003locally} identified a problem that not all all-inlier samples are ``good'', i.e.\ lead to a model accurate enough to distinguish all inliers, e.g.\ due to poor conditioning of the selected random all-inlier sample. They address the problem by introducing the locally optimized RANSAC (LO-RANSAC) that augments the original approach with a local optimization step applied to the \textit{so-far-the-best} model. 
In the original paper~\cite{chum2003locally}, local optimization is implemented as an iterated least squares re-fitting with a shrinking inlier-outlier threshold inside an inner RANSAC applied only to the inliers of the current model. In the reported experiments, LO-RANSAC outperforms standard RANSAC in both accuracy and the required number of iterations. The number of LO runs is close to the logarithm of the number of verifications, and it does not create a significant overhead in the processing time in most of the cases tested.
However, it was shown by Lebeda et al.~\cite{lebeda2012fixing} that for models with high inlier counts the local optimization step becomes a computational bottleneck due to the iterated least squares model fitting. This is addressed by using a $7m$-sized subset of the inliers in each LO step, where $m$ is the size of a minimum sample; the factor of $7$ was set by exhaustive experimentation.
The idea of local optimization has been included in state-of-the-art RANSAC approaches like USAC~\cite{raguram2013usac}. Nevertheless, the LO procedure remains ad hoc, complex and requires multiple parameters.

In this paper, we combine two strands of research to obtain a state-of-the-art RANSAC. 
In the large body of RANSAC-related literature, the inlier-outlier decision has always been a function of the distance to the model, done individually for each data point. Yet both inliers and outliers are spatially coherent, a point near an outlier or inlier is more likely to be an outlier or inlier respectively. 
Spatial coherence, leading to the Potts model~\cite{boykov1998markov}, has been exploited in many vision problems, for instance, in segmentation~\cite{zabih2004spatially}, multi-model fitting~\cite{isack2012energy,pham2014interacting} or sampling~\cite{nasuto2002napsac}. 
In RANSAC techniques, it has only been used to improve efficiency of sampling in NAPSAC~\cite{nasuto2002napsac}.
%
%
It is computationally prohibitive to formulate the model verification in RANSAC as a graph-cut problem. But when applied as the LO step in \cite{chum2003locally} just to the \textit{so-far-the-best} model, the number of graph-cuts is only the logarithm of the number of sampled and verified models, and can be achieved in real-time.

The proposed  method, called Graph-Cut RANSAC (GC-RANSAC), is a locally optimized RANSAC alternating graph-cut and model re-fitting as the LO step. GC-RANSAC is superior to LO-RANSAC in a number of aspects.
First,  it is capable of exploiting spatial coherence of inliers and outliers. The LO step is conceptually a simple, easy to implement, globally optimal and computationally efficient graph-cut with only a few intuitive and learnable parameters unlike the ad hoc, iterative and complex LO steps~\cite{chum2003locally}. 
Third, we show experimentally that GC-RANSAC outperforms LO-RANSAC and its recent variants in both accuracy and the required number of iterations on a wide range of publicly available datasets. On many problems, it is faster than the competitors in terms of the wall-clock time. 
Finally, we were surprised to observe that GC-RANSAC terminates {\it before} the theoretically expected number of iterations. The reason is that the local optimization that takes spatial proximity into account is often capable of converging to a ``good'' model even when starting from a sample that is not all-inlier, i.e.\ it contains outliers.

PEARL~\cite{isack2012energy} introduced pair-wise energy to geometric model fitting. However, it cannot be used for problems solved by RANSAC -- in PEARL, the user has to manually set the number of hypotheses tested to the worst-case, i.e.\ corresponding to the lowest inlier ratio possible. The $\alpha$-expansion step just in the first iteration of PEARL executes a graph-cut as many times as the number of hypotheses tested. The number is calculated from the worst-case scenario and is typically orders of magnitude higher than the number of iterations determined by the RANSAC adaptive termination criterion.
Moreover, in GC-RANSAC, applying the local optimization to only the \textit{so-far-the-best} models ensures that the graph-cut is executed only very few times, paying only a small penalty.

%


\section{Local Optimization and Spatial Coherence}

In this section, we formulate the inlier selection of RANSAC as an energy minimization considering point-to-point proximity. The proposed local optimization is seen as an iterative energy minimization of a binary labeling (outlier -- 0 and inlier -- 1). 
For the sake of simplicity, we start from the original RANSAC scheme and then formulate the maximum likelihood estimation as an energy minimization. The term considering the spatial coherence will be included into the energy.

\subsection{Formulation as Energy Minimization}  

Suppose that a point set $\mathcal{P} \subseteq \mathbb{R}^n$ ($n > 0$), a model represented by a parameter vector $\theta \in \mathbb{R}^m$ ($m > 0$) and a distance function $\phi: \mathcal{P} \times \mathbb{R}^m \rightarrow \mathbb{R}$ measuring the point-to-model assignment cost are given.

For the standard RANSAC scheme which applies a top-hat fitness function ($1$ -- close, $0$ -- far), the implied unary energy is as follows: 
\begin{equation*}
	E_{\{0;1\}}(L) = \sum_{p \in \mathcal{P}} ||L_p||_{\{0;1\}},
\end{equation*}
where 
\begin{equation*}
	||L_p||_{\{0;1\}} = 
    \begin{cases}
		0 & \text{if } (L_p = 1 \wedge \phi(p, \theta) < \epsilon) \text{ } \vee \\    
		 & \text{\phantom{xx}} (L_p = 0 \wedge \phi(p, \theta) \geq \epsilon) \\    
        1 & \text{otherwise.}
	\end{cases}
\end{equation*}
Parameter $L \in \{0, 1\}^{|\mathcal{P}|}$ is a labeling, ignored in standard RANSAC, $L_p \in L$ is the label of point $p \in \mathcal{P}$, $|\mathcal{P}|$ is the number of points, and $\epsilon$ is the inlier-outlier threshold. Using energy $E_{\{0,1\}}$ we get the same result as RANSAC since it does not penalize only two cases: (i) when $p$ is labeled inlier and it is closer to the model than the threshold, or (ii) when $p$ is labeled outlier and it is farer from the model than $\epsilon$. This is exactly what RANSAC does. 

Since the publication of RANSAC, several papers discussed, e.g.\ \cite{lebeda2012fixing}, replacing the $\{0, 1\}$ loss with a kernel function $K: \mathbb{R} \times \mathbb{R} \rightarrow [0,1]$, e.g.\ the Gaussian-kernel.  Such choice is close to maximum likelihood estimation as proposed in MLESAC~\cite{torr2000mlesac}. This improves the accuracy and reduces the sensitivity to threshold $\epsilon$. Unary term $E_\text{K}$ exploiting this continuous loss is as follows:
$ 
	E_{\text{K}}(L) = \sum_{p \in \mathcal{P}} ||L_p||_\text{K},
$ 
where 
\begin{equation}
	\label{eq:unary}
	||L_p||_{\text{K}} = 
    \begin{cases}
		1 - K(\phi(p, \theta), \epsilon) & \text{if } L_p = 1 \text{ } \\      
        K(\phi(p, \theta), \epsilon) & \text{if } L_p = 0 \text{ }
	\end{cases}
\end{equation}
and
\begin{equation}
	\label{eq:kernel_function}
	K(\delta, \epsilon) = e^{-\frac{\delta^2}{2\epsilon^2}},
\end{equation}
which equals to one if the distance is zero. In GC-RANSAC, we use $E_\text{K}$ as the unary energy term in the graph-cut-based verification.


\subsection{Spatial Coherence}  

Benefiting from a binary labeling energy minimization, additional energy terms, e.g.\ to consider spatial coherence of the points, can be included yet keep the problem \textit{solvable efficiently and globally via the graph-cut algorithm}. 

Considering point proximity is a well-known approach for sampling~\cite{nasuto2002napsac} or multi-model fitting~\cite{isack2012energy,pham2014interacting,barath2016multi}. 
To the best of our knowledge, there is no paper exploiting it in the local optimization step of methods like LO-RANSAC. 
Applying the Potts model which penalizes all neighbors having different labels would be a justifiable choice to be the pair-wise energy term. The problem arises when the data contains significantly more outliers, probably close to desired model, than inliers. In that case, penalizing differently labeled neighbors using the same penalty for all classes many times leads to the domination of outliers forcing all inliers to be labeled outlier. To overcome this problem, we modified the Potts model to use different penalty for each neighboring point pair on the basis of their inlier probability. The proposed pair-wise energy term is  
\begin{equation}
	\label{eq:pairwise}
	E_\text{S}(L) = \sum_{(p,q) \in \mathcal{A}}
    \begin{cases}
		1 & \text{if } L_p \neq L_q \\      
        \frac{1}{2} (K_{p} + K_{q}) & \text{if } L_p = L_q = 0 \\
        1 - \frac{1}{2} (K_{p} + K_{q}) & \text{if } L_p = L_q = 1 \\
	\end{cases},
\end{equation}
where $K_{p} = K(\phi(p, \theta), \epsilon)$, $K_{q} = K(\phi(q, \theta), \epsilon)$ and $(p,q)$ is an edge of neighborhood graph $\mathcal{A}$ between points $p$ and $q$. In $E_\text{S}$, if both points labeled outlier the penalty is $\frac{1}{2} (K_{p} + K_{q})$ thus ``rewarding'' label $0$ if the neighboring points are far from the model. The penalty of considering a point as inlier is $1 - \frac{1}{2} (K_{p} + K_{q})$ which rewards the label if the points are close to the model.

The proposed overall energy measuring the fitness of points to a model and considering spatial coherence is $E(L) = E_{\text{K}}(L) + \lambda E_\text{S}(L)$, where $\lambda$ is a parameter balancing the terms. 
The globally optimal labeling $L^* = \arg \min_{L} E(L)$ can easily be determined in polynomial time using graph-cut algorithm. 

\section{GC-RANSAC}

In this section, we include the proposed energy minimization-based local optimization into RANSAC. Benefiting from this approach, the LO step is getting simpler and cleaner than that of LO-RANSAC. 

The main algorithm is shown in Alg.\ref{alg:gcransac}. The first step is the determination of neighborhood graph $\mathcal{A}$ for which we use a sphere with a predefined radius $r$ -- this is a parameter of the algorithm -- and Fast Approximate Nearest Neighbors algorithm~\cite{muja2009fast}. In Alg.~\ref{alg:gcransac}, function $H$ is as follows~\cite{fischler1981random}:  
\begin{equation}   
	H(|L^*|, \mu) = \frac{\log(\mu)}{\log(1 - P_I)},
    \label{eq:H}
\end{equation}
where $P_I = \binom{|L^*|}{m} / \binom{|P|}{m}.$
%
%
It calculates the required iteration number of RANSAC on the basis of desired probability $\mu$, the size of the required minimal point set $m$ and the inlier number $|L^*|$ regarding to the current \textit{so-far-the-best} model. Note that norm $| \cdot |$ applied to the labeling counts the inliers.  

Every $k$th iteration draws a minimal sample using a sampling strategy, e.g.\ PROSAC~\cite{chum2005matching}, then computes the parameters $\theta_k$ of the implied model and its support
\begin{equation}    
	w_k = \sum_{p \in \mathcal{P}} \text{K}(\phi(p, \theta_k), \epsilon)
    \label{eq:support}
\end{equation}
w.r.t.\ the data points, where function $K$ is a Gaussian-kernel as proposed in Eq.~\ref{eq:kernel_function}. If $w_k$ is higher than that of the \textit{so-far-the-best} model $w^*$, this model is considered the new \textit{so-far-the-best}, all parameters are updated, i.e.\ the labeling, model parameters and support, and local optimization is applied if needed. Note that the application criterion of the local optimization step is discussed later. 


The proposed local optimization is written in Alg.~\ref{alg:lo_optimization}. The main iteration can be considered as a grab-cut-like~\cite{rother2004grabcut} alternation consisting of two major steps: (i) graph-cut and (ii) model re-fitting. The construction of problem graph $G$ using unary and pair-wise terms Eqs.~\ref{eq:unary},~\ref{eq:pairwise} is shown in Alg.~\ref{alg:graph_construction}. Functions AddTerm1 and AddTerm2 are discussed in~\cite{kolmogorov2004energy} in depth. Graph-cut is applied to $G$ determining the optimal labeling $L$ which considers the spatial coherence of the points and their distances from the \textit{so-far-the-best} model. Model parameters $\theta$ are computed using a $7m$-sized random subset of the inliers in $L$, thus speeding up the process, similarly to~\cite{lebeda2012fixing} does, where $m$ is the size of a minimal sample, e.g.\ $m = 2$ for lines. Note that $7m$ is set by exhaustive experimentation in~\cite{lebeda2012fixing} and this value also suited for us. Finally, the support $w$ of $\theta$ is computed and the \textit{so-far-the-best} model is updated if the new one has higher support, otherwise the process terminates. 
After the main algorithm, a local optimization step is performed if it has not been yet applied to the obtained \textit{so-far-the-best} model. Then the model parameters are re-estimated using the whole inlier set similarly to plain RANSAC does. 

Remark: Adding to the local optimization step a RANSAC-like procedure selecting $7m$-size samples is straightforward. In our experiments, it had a high computational overhead without adding significantly to accuracy.

\begin{algorithm}
\begin{algorithmic}[1]
	\Statex{\hspace{-1.0em}\textbf{Input:} $\mathcal{P}$ -- data points; $r$ -- sphere radius, $\epsilon$ -- threshold}
	\Statex{\hspace{-1.0em}\phantom{xxxxxx} $\epsilon_{\text{conf}}$ -- LO application threshold, $\mu$ -- confidence;}
    \Statex{\hspace{-1.0em}\textbf{Output:} $\theta$ - model parameters; $L$ -- labeling }
   	\Statex{}
    \State{$w^*, n_{LO} \leftarrow 0, 0$.}
    \State{$\mathcal{A} \leftarrow $ Build neighborhood-graph using $r$.}
    \For{k = $1 \to H(|L^*|, \mu)$} \Comment Eq.~\ref{eq:H}
    	\State{ $S_k \leftarrow $ Draw a minimal sample.}
    	\State{ $\theta_k \leftarrow $ Estimate a model using $S_k$.}
    	\State{ $w_k \leftarrow $ Compute the support of $\theta_k$.}\Comment Eq.~\ref{eq:support}
        \If {$w_k > w^*$}
        	\State{$\theta^*, L^*, w^* \leftarrow \theta_k, L_k, w_k$}
        	\If {ApplyLocalOptimization($\epsilon_{\text{conf}})$}
		        \State{$\theta_{LO}, L_{LO}, w_{LO} \leftarrow $ Local opt. } \Comment Alg.~\ref{alg:lo_optimization}
		        \State{$n_{LO} \leftarrow n_{LO} + 1$. }
	 	    	\If {$w_{LO} > w^*$}
        			\State{$\theta^*, L^*, w^* \leftarrow \theta_{LO}, L_{LO}, w_{LO}$}
                \EndIf         
        	\EndIf  	
        \EndIf
    \EndFor
    \If {$n_{LO} = 0$}
		\State{$\theta^*, L^*, w^* \leftarrow $ Local opt. } \Comment Alg.~\ref{alg:lo_optimization}
    \EndIf
    \State{$\theta^* \leftarrow $ least squares model fitting using $L^*$.}
\end{algorithmic}
\caption{\bf The GC-RANSAC Algorithm.}
\label{alg:gcransac}
\end{algorithm}

\begin{algorithm}
\begin{algorithmic}[1]
	\Statex{\hspace{-1.0em}\textbf{Input:} $\mathcal{P}$ -- data points, $L^*$ -- labeling, }
	\Statex{\hspace{-1.0em}\phantom{xxxxxx} $w^*$ -- support, $\theta^*$ -- model;}
    \Statex{\hspace{-1.0em}\textbf{Output:} $L_{LO}^*$ -- labeling, $w_{LO}^*$ -- support, $\theta_{LO}^*$ -- model; }
   	\Statex{}
    \State{$w_{LO}^*, L_{LO}^*, \theta_{LO}^*, changed \leftarrow w^*, L^*, \theta^*, 1$.}
    \While{$changed$}
	    \State{ $G \leftarrow $ Build the problem graph. } \Comment Alg.~\ref{alg:graph_construction}
    	\State{ $L \leftarrow $ Apply graph-cut to $G$. }
    	\State{ $I_{7m} \leftarrow $ Select a $7m$-sized random inlier set. }
    	\State{ $\theta \leftarrow $ Fit a model using labeling $I_{7m}$. }
        \State{ $w \leftarrow $ Compute the support of $\theta$. }
        \State{ $changed \leftarrow 0$. }
        \If {$w > w_{LO}^*$}
        	\State{$\theta_{LO}^*, L_{LO}^*, w_{LO}^*, changed \leftarrow \theta, L, w, 1$.}  
        \EndIf
    \EndWhile
\end{algorithmic}
\caption{\bf Local optimization.}
\label{alg:lo_optimization}
\end{algorithm}

\begin{algorithm}
\begin{algorithmic}[1]
	\Statex{\hspace{-1.0em}\textbf{Input:} $\mathcal{P}$ -- data points, $\mathcal{A}$ -- neighborhood-graph }
	\Statex{\hspace{-1.0em}\phantom{xxxxxx} $\theta$ -- model parameters, $\theta^*$ -- model;}
    \Statex{\hspace{-1.0em}\textbf{Output:} $G$ -- problem graph; }
   	\Statex{}
    \State{$G \leftarrow $ EmptyGraph().}
    \For{$p \in \mathcal{P}$}
    	\State{$c_0, c_1 \leftarrow \text{K}(\phi(p, \theta), 1 - \text{K}(\phi(p, \theta), \epsilon)$}
    	\State{$G \leftarrow $ AddTerm1($G$, $p$, $c_0$, $c_1$).}
    \EndFor
    
    \For{$(p,q) \in \mathcal{A}$}
    	\State{$c_{01}, c_{10} \leftarrow 1, 1$.}
   	 	\State{$c_{00} \leftarrow 0.5 (\text{K}(\phi(q, \theta) + \text{K}(\phi(p, \theta))$.}
    	\State{$c_{11} \leftarrow 1 - 0.5 (\text{K}(\phi(q, \theta) + \text{K}(\phi(p, \theta))$.}
    	\State{$G \leftarrow $ AddTerm2($G$, $p$, $q$, $c_{00}$, $c_{01}$, $c_{10}$, $c_{11}$).}
    \EndFor
\end{algorithmic}
\caption{\bf Problem Graph Construction.}
\label{alg:graph_construction}
\end{algorithm}

\textbf{The criterion for applying the LO step} was proposed to be: (i) the model is so-far-the-best and (ii) after a user-defined iteration limit, in~\cite{lebeda2012fixing}.  However, in our experiments, this approach still spends significant time on optimizing models which are not promising enough. We introduce a simple heuristics for replacing the iteration limit with a data driven strategy which allows to apply LO only a few times without deterioration in accuracy. 

As it is well-known for RANSAC, the required iteration number $k$, w.r.t.\ the inlier ratio $\eta$, sample size $m$ and confidence $\mu$, is calculated as $k = \log (1 - \mu) / \log (1 - \eta^m)$.
Re-arranging this formula to $\mu$ leads to equation 
$
	\mu = 1 - 10^{k \log (1 - \eta^m)}
$ 
which determines the confidence of finding the desired model in the $k$th iteration if the inlier ratio is $\eta$. 

Suppose that the algorithm finds a new so-far-the-best model with inlier ratio $\eta_2$ in the $k_2$th iteration, whilst the previous best model was found in the $k_1$th iteration with inlier ratio $\eta_1$ ($k_2 > k_1$, $\eta_2 > \eta_1$). The ratio of the confidences $\mu_{12}$ in those two models is calculated as follows:
\begin{equation}
	\mu_{12} = \frac{\mu_2}{\mu_1} = \frac{1 - 10^{k_2 \log (1 - \eta_2^m)}}{1 - 10^{k_1 \log (1 - \eta_1^m)}}.
    \label{eq:criterion}
\end{equation}
In experiments, we observed that a model that leads to termination if optimized often shows a significant increase in the confidence. Replacing the parameter blocking LO in the first $k$ iterations, we adopt a criterion $q_{12} > \epsilon_{\text{conf}}$, where $\epsilon_{\text{conf}}$ is a user-defined parameter determining  a significant increase. 

\section{Experimental Results}

In this section, GC-RANSAC is validated both on synthesized and publicly available real world data and compared with plain RANSAC~\cite{fischler1981random}, LO-RANSAC~\cite{chum2003locally}, LO$^+$-RANSAC, LO'-RANSAC~\cite{lebeda2012fixing}, and EP-RANSAC~\cite{leexact2017}. The parameter setting is reported in Table~\ref{tab:parameter_setup}. For EP-RANSAC\footnote{The Matlab source is available at \url{http://cs.adelaide.edu.au/~huu/publication/exact_penalty/}}, we tuned the threshold parameter to achieve the lowest mean error and the other parameters were set to the values reported by the authors. Note that the comparison of the processing time with this method is affected by the availability of a Matlab implementation only. 
All methods apply PROSAC~\cite{chum2005matching} sampling and encapsulates the \textit{point-to-model} distance, e.g.\ re-projection error for homographies, with a Gaussian-kernel using $\epsilon = 0.31$, which is set by an exhaustive search. EP-RANSAC uses inlier maximization strategy since its cost function cannot be replaced straightforwardly. The radius of the sphere to determine neighboring points is $20$ pixels and it is applied to the concatenated $4D$ coordinates of the correspondences. Parameter $\lambda$ for GC-RANSAC was set to $0.1$ and $\epsilon_{\text{conf}} = 10$. 

\begin{table}
\center
\caption{ Setting for the tests. Outlier threshold ($\epsilon$), radius used for proximity computation ($r$), weight of the pair-wise term ($\lambda$), and the threshold of the confidence change ($\epsilon_{\text{conf}}$).}
	\begin{tabular}{| c | c | c | c |  }
    \hline 
 	 	$\epsilon$ & $r$ & $\lambda$ & $\epsilon_{\text{conf}}$ \\ 
    \hline     
   		0.31 & 20 px & 0.10 & 10 \\ 
    \hline     
\end{tabular}
\label{tab:parameter_setup}
\end{table}

\begin{figure}
  \centering
  \begin{subfigure}[b]{0.45\columnwidth}
 	 	\centering
  		\includegraphics[width=1.0\columnwidth]{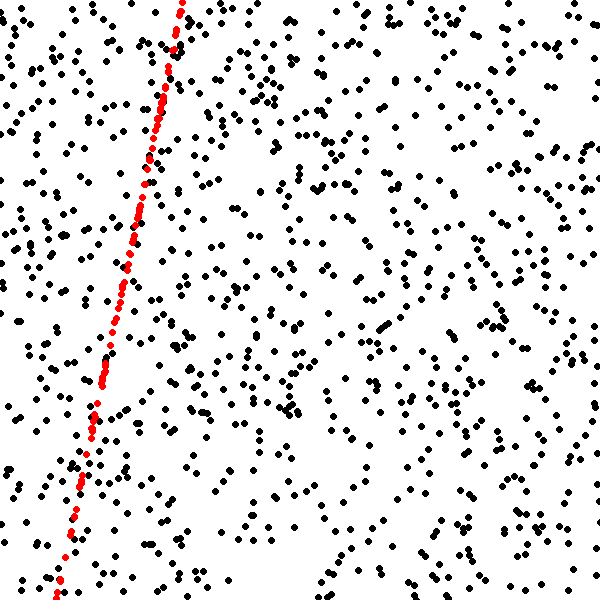}
        \caption{}
    	\label{fig:straight_line}
  \end{subfigure}
  \begin{subfigure}[b]{0.45\columnwidth}
 	 	\centering
  		\includegraphics[width=1.0\columnwidth]{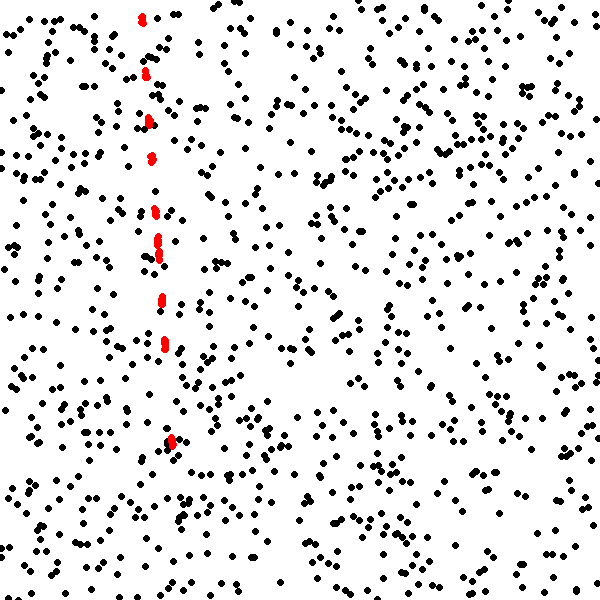}
        \caption{}
  		\label{fig:dashed_line}
  \end{subfigure}
  \caption{An example input for (a) straight and (b) dashed lines. The 1000 black points are outliers, the 100 red ones are inliers. \textit{Best viewed in color.} }
\end{figure}

\paragraph{Synthetic Tests on 2D Lines.} 
To compare GC-RANSAC with the state-of-the-art in a fully controlled environment, we chose two simple tests: detection of a 2D straight or dashed line. For each trial, a $600 \times 600$ window and a random line was generated in its implicit form, sampled at $100$ locations and zero-mean Gaussian-noise with $\sigma$ standard deviation was added to the coordinates. For a straight line, the points were generated using uniform distribution (see Fig.~\ref{fig:straight_line}). For a dashed line, $10$ knots were put randomly into the window, then the line is sampled at $10$ locations with uniform distribution around each knot, at most $10$ pixels far (see Fig.~\ref{fig:dashed_line}). Finally, $k$ outliers were added to the scene. $1000$ tests were performed on every noise level. 

\begin{figure}
  \centering
  \begin{subfigure}[b]{0.49\columnwidth}
 	 	\centering
  		\includegraphics[width=1.0\columnwidth]{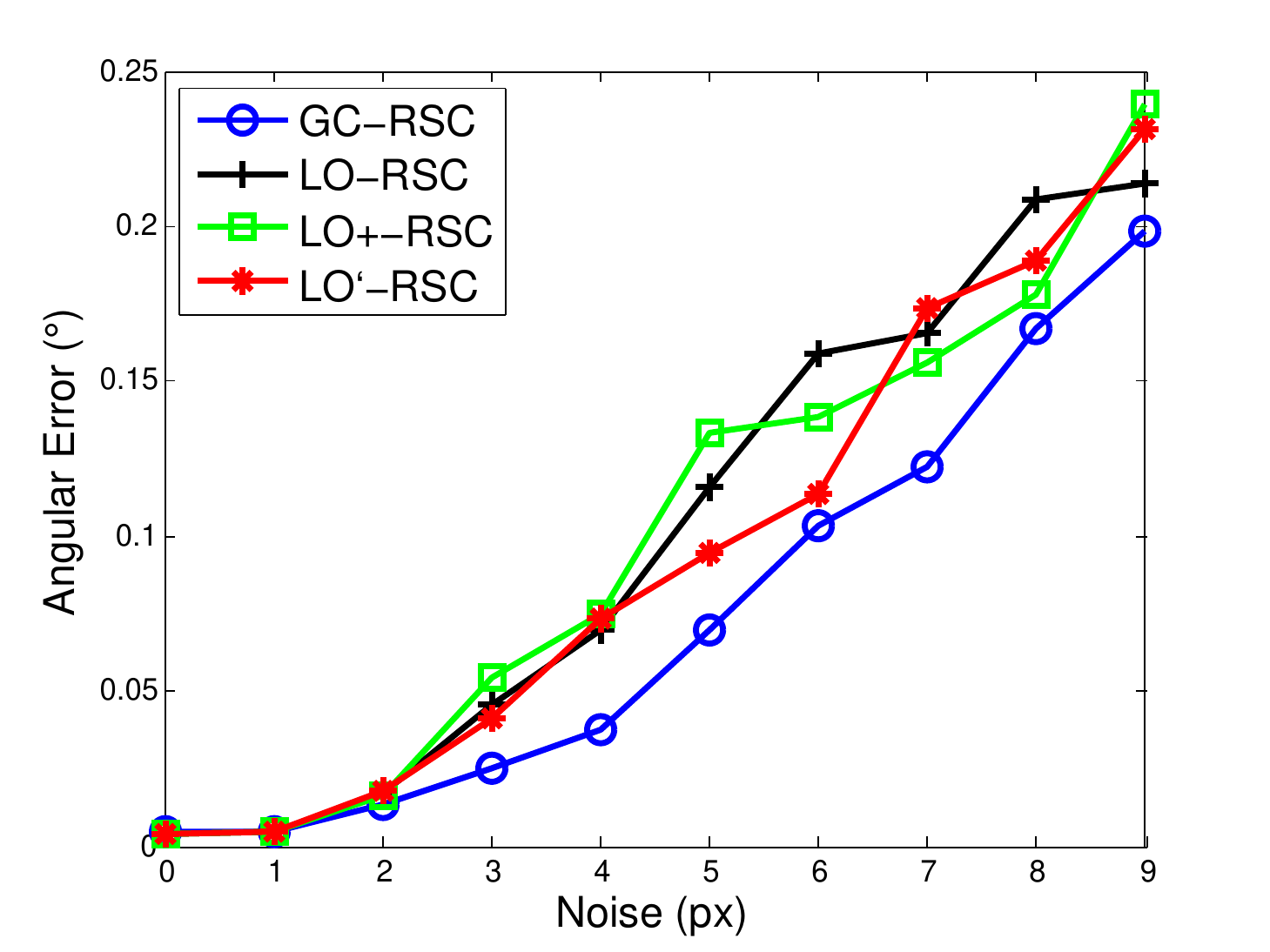}
        \caption{}
  		\label{fig:line_plots_a}
  \end{subfigure}
  \begin{subfigure}[b]{0.49\columnwidth}
 	 	\centering
  		\includegraphics[width=1.0\columnwidth]{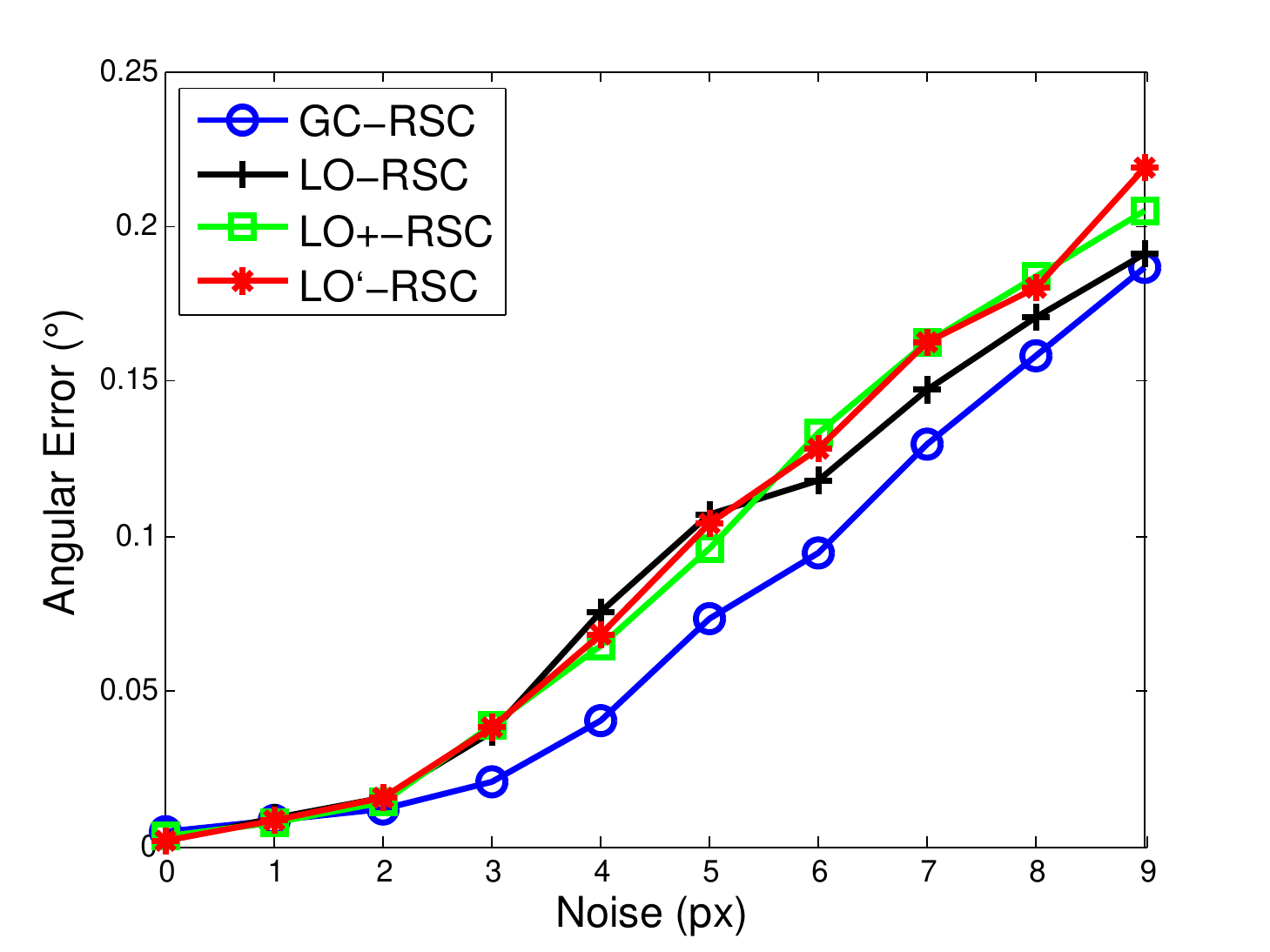}
        \caption{}
 		\label{fig:line_plots_b}
  \end{subfigure}\\
  \begin{subfigure}[b]{0.49\columnwidth}
 	 	\centering
  		\includegraphics[width=1.0\columnwidth]{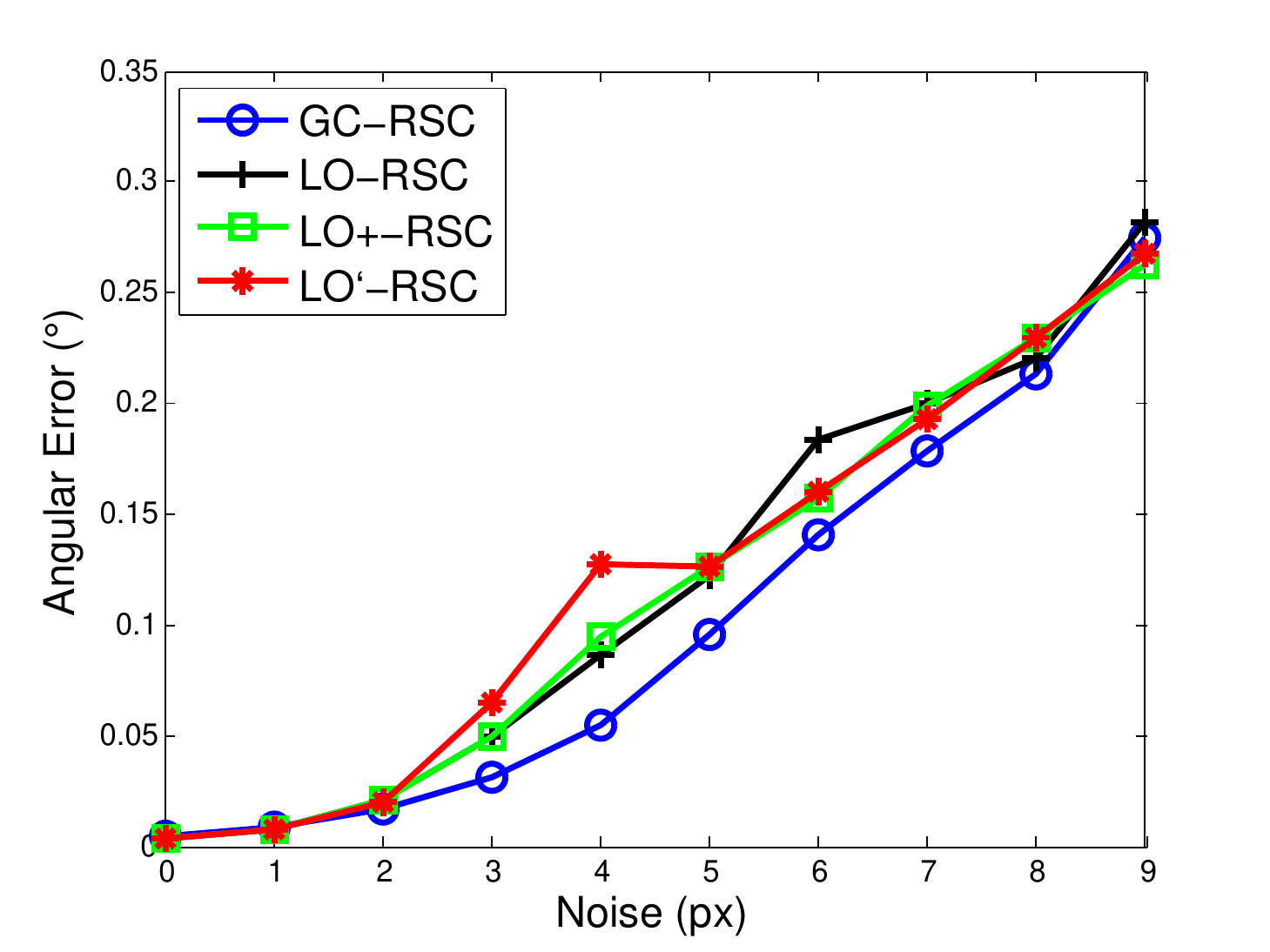}
        \caption{}
  		\label{fig:line_plots_c}
  \end{subfigure}
  \begin{subfigure}[b]{0.49\columnwidth}
 	 	\centering
  		\includegraphics[width=1.0\columnwidth]{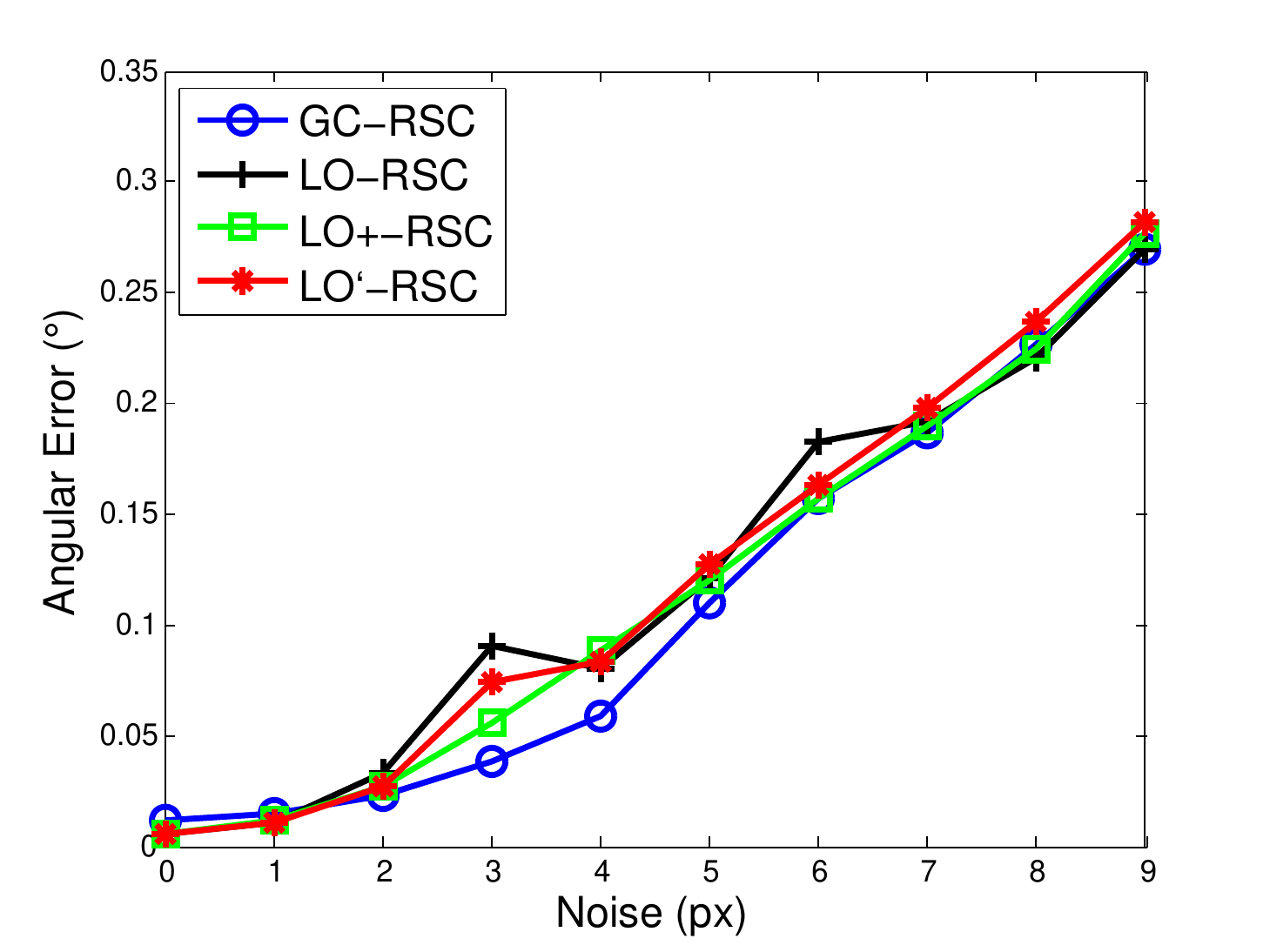}
        \caption{}
  		\label{fig:line_plots_d}
  \end{subfigure}
        \caption{The mean angular error (in degrees) of the obtained 2D lines plotted as the function of noise $\sigma$ (in pixels). On each noise $\sigma$, $1000$ runs were performed. The line type and outlier number is (a) straight line, $100\%$, (b) straight line, $500\%$ (c) dashed line, $100\%$ and (c) dashed line, $500\%$. 
        }
  \label{fig:line_plots}
\end{figure}

Fig.~\ref{fig:line_plots} shows the mean angular error (in degrees) plotted as the function of the noise $\sigma$. The first and second rows report the results of the straight and dashed line cases. For the two columns, $100$ and $500$ outliers were added, respectively. 
According to Fig.~\ref{fig:line_plots}, \textit{GC-RANSAC obtains more accurate lines than the competitor algorithms}.

\begin{table}
\center
	\caption{ Percentage of ``not-all-inlier'' minimal samples leading to the correct solution during line ($\mathbf{L}$) and fundamental matrix ($\mathbf{F}$) fitting. For lines, the average over $1000$ runs on three different outlier percentage (100\%, 500\%, 1000\%) and noise levels $0.0 - 9.0$ px is reported, thus $15000$ runs were performed. For $\mathbf{F}$, the mean of $1000$ runs on the {\fontfamily{cmtt}\selectfont AdelaideRMF} dataset is shown.}
  	\resizebox{0.69\linewidth}{!}{\begin{tabular}{| r | c | c | c | c |  }
    \hline 
 	 	& LO & LO$^+$ & LO' & \textbf{GC} \\ 
    \hline     
   		\multirow{1}{*}{$\mathbf{L}$} & \phantom{x}6\% & \phantom{x}5\% & \phantom{x}4\% & \textbf{15\%} \\
   		\multirow{1}{*}{$\mathbf{F}$} & 29\% & 30\% & 24\% & \textbf{32\%}\\
    \hline     
\end{tabular}}
\label{tab:not_all_inl_comparison}
\end{table}

\begin{figure}
  \centering
  \begin{subfigure}[b]{0.99\columnwidth}
 	 	\centering
  		\includegraphics[width=0.86\columnwidth]{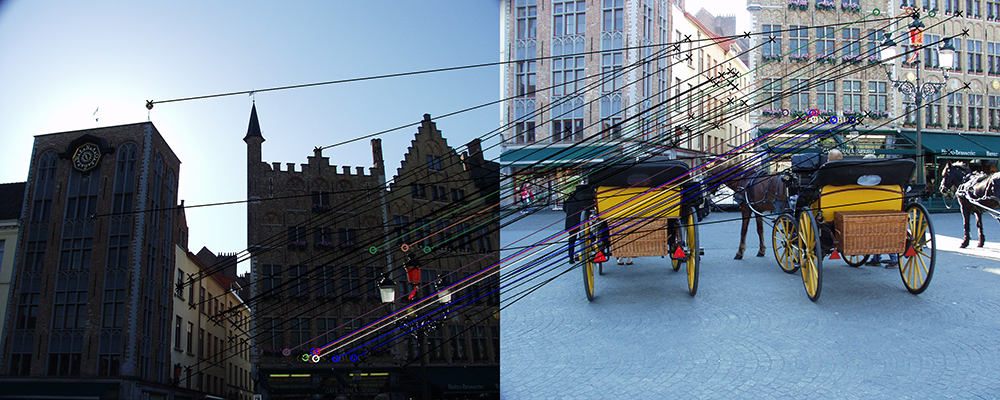}
        \caption{Homography; {\fontfamily{cmtt}\selectfont homogr} dataset}
  \end{subfigure}
  \begin{subfigure}[b]{0.99\columnwidth}
 	 	\centering
  		\includegraphics[width=0.86\columnwidth]{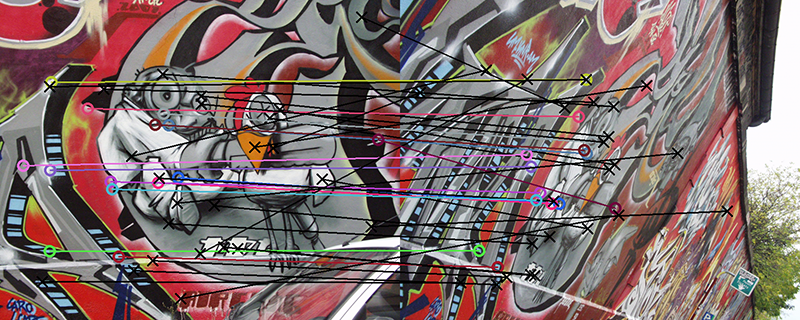}
        \caption{Homography; {\fontfamily{cmtt}\selectfont EVD} dataset}
  \end{subfigure}
  \begin{subfigure}[b]{0.99\columnwidth}
 	 	\centering
  		\includegraphics[width=0.86\columnwidth]{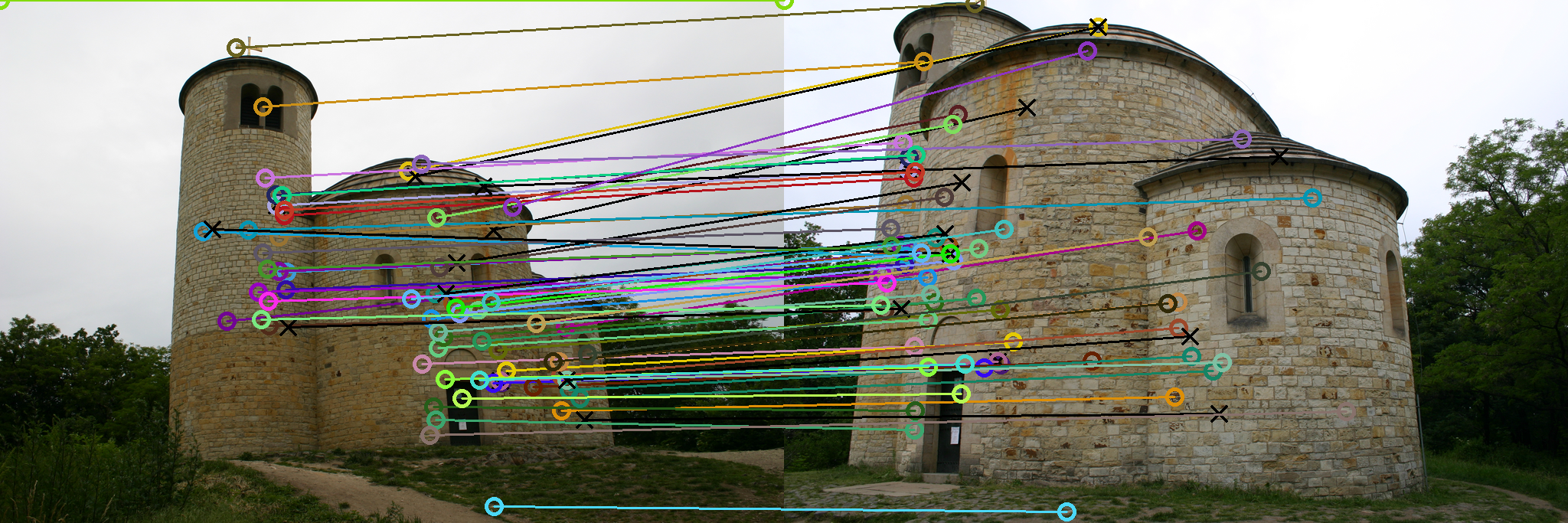}
        \caption{Fundamental matrix; {\fontfamily{cmtt}\selectfont kusvod2} dataset}
  \end{subfigure}
  \begin{subfigure}[b]{0.99\columnwidth}
 	 	\centering
  		\includegraphics[width=0.86\columnwidth]{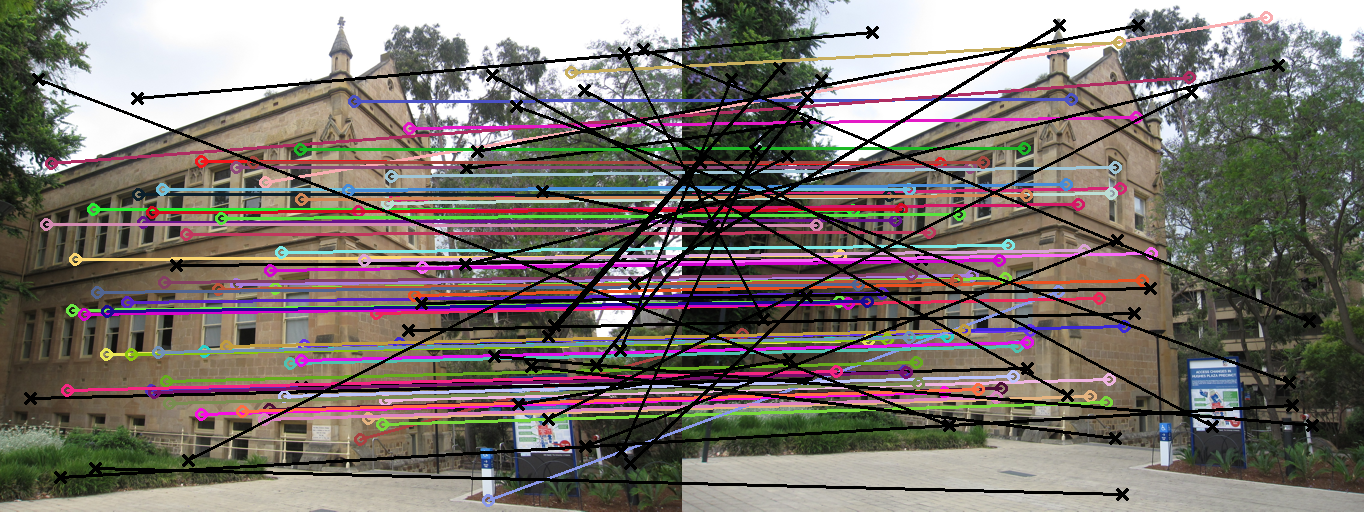}
        \caption{Fundamental matrix; {\fontfamily{cmtt}\selectfont AdelaideRMF} dataset}
  \end{subfigure}
  \begin{subfigure}[b]{0.99\columnwidth}
 	 	\centering
  		\includegraphics[width=0.86\columnwidth]{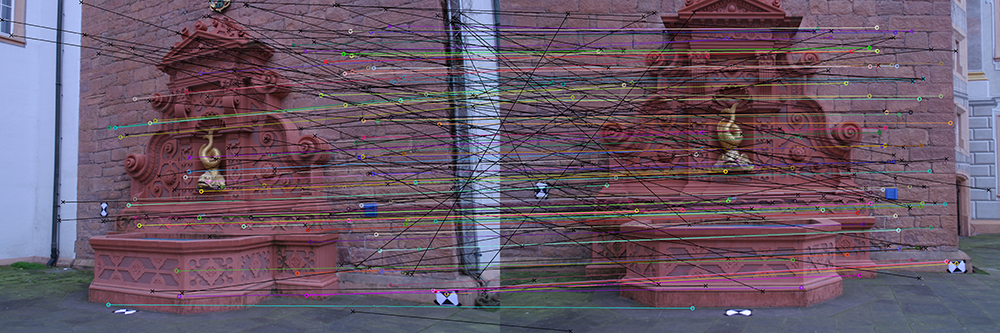}
        \caption{Essential matrix; {\fontfamily{cmtt}\selectfont Strecha} dataset}
  \end{subfigure}
  \begin{subfigure}[b]{0.99\columnwidth}
 	 	\centering
  		\includegraphics[width=0.86\columnwidth]{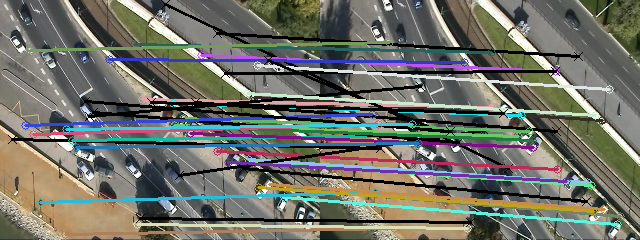}
        \caption{Affine transformation; {\fontfamily{cmtt}\selectfont SZTAKI} dataset}
  \end{subfigure}
  \caption{ Results of GC-RANSAC on example pairs from each dataset and problem. Correspondences are drawn by lines and circles, outliers by black lines and crosses, every third correspondence is drawn. }
  \label{fig:example_images_hom}
\end{figure}

\paragraph{Estimation of Fundamental Matrix.}  
Evaluating the performance of GC-RANSAC on fundamental matrix estimation we used {\fontfamily{cmtt}\selectfont kusvod2} (24 pairs)\footnote{\url{http://cmp.felk.cvut.cz/data/geometry2view/}}, {\fontfamily{cmtt}\selectfont Multi-H}\footnote{\url{http://web.eee.sztaki.hu/~dbarath/}} (5 pairs), and  {\fontfamily{cmtt}\selectfont AdelaideRMF}\footnote{\url{cs.adelaide.edu.au/~hwong/doku.php?id=data}} (19 pairs) datasets (see Fig.~\ref{fig:example_images_hom} for examples). {\fontfamily{cmtt}\selectfont Kusvod2} consists of 24 image pairs of different sizes with point correspondences and fundamental matrices estimated using manually selected inliers. {\fontfamily{cmtt}\selectfont AdelaideRMF} and {\fontfamily{cmtt}\selectfont Multi-H} consist a total of 24 image pairs with point correspondences, each assigned manually to a homography (or the outlier class). For them, all points which are assigned to a homography were considered as inliers and others as outliers. On total, the proposed method was tested on 48 image pairs from three publicly available datasets for fundamental matrix estimation.  
All methods applied the 7-point method~\cite{hartley2003multiple} to estimate $\textbf{F}$, thus drawing minimal sets of size seven in each RANSAC iteration. For the model re-estimation from a non-minimal sample in the LO step, the normalized 8-point algorithm~\cite{hartley1997defense} is used. Note that all fundamental matrices were discarded for which the \textit{oriented} epipolar constraint~\cite{chum2004epipolar} did not hold.

The first three blocks of Table~\ref{tab:dataset_comparison}, each consisting of four rows, report the quality of the epipolar geometry estimation on each dataset as the average of 1000 runs on every image pair. The first two columns show the name of the tests and the investigated properties: 
\textbf{(1)} LO: the number of applied local optimization steps (graph-cut steps are shown in brackets). \textbf{(2)} $\mathcal{E}$ is the geometric error (in pixels) of the obtained model w.r.t.\ the manually annotated inliers. For fundamental matrices and homographies, it is defined as the average Sampson distance and re-projection error, respectively. For essential matrices, it is the mean Sampson distance of the implied fundamental matrix and the correspondences. 
\textbf{(3)} $\mathcal{T}$ is the mean processing time in milliseconds. 
\textbf{(4)} $\mathcal{S}$ is the average number of minimal samples have to be drawn until convergence, basically, the number of RANSAC iterations. 

It can be clearly seen that for fundamental matrix estimation GC-RANSAC \textit{always obtains the most accurate model} using less samples than the competitive methods. 

\paragraph{Estimation of Homography.}  
In order to test homography estimation we downloaded {\fontfamily{cmtt}\selectfont homogr}\footnote{\url{http://cmp.felk.cvut.cz/data/geometry2view/}} (16 pairs) and {\fontfamily{cmtt}\selectfont EVD}\footnote{\url{http://cmp.felk.cvut.cz/wbs/}} (15 pairs) datasets (see Fig.~\ref{fig:example_images_hom} for examples). Each consists of image pairs of different sizes from $329 \times 278$ up to $1712 \times 1712$ with point correspondences and manually selected inliers -- correctly matched point pairs.   
{\fontfamily{cmtt}\selectfont Homogr} dataset consists of short baseline stereo pairs, whilst the pairs of {\fontfamily{cmtt}\selectfont EVD} undergo an extreme view change, i.e.\ wide baseline. 
All methods apply the normalized four-point algorithm~\cite{hartley2003multiple} for homography estimation both in the model generation and local optimization steps. Therefore, each minimal sample consists of four correspondences. 

The $4$th and $5$th blocks of Fig.~\ref{tab:dataset_comparison} show the mean results computed using all the image pairs of each dataset. 
It can be seen that GC-RANSAC obtains the most accurate models for all but one, i.e.\ {\fontfamily{cmtt}\selectfont EVD} dataset with time limit, test cases. 

\paragraph{Estimation of Essential Matrix.}  
To estimate essential matrices, we used the {\fontfamily{cmtt}\selectfont strecha} dataset~\cite{strecha2004wide} consisting of image sequences of buildings. 
All image sizes are $3072 \times 2048$. The ground truth projection matrices are provided. 
The methods were applied to all possible image pairs in each sequence. The SIFT detector~\cite{lowe1999object} was used to obtain correspondences. For each image pair, a reference point set with ground truth inliers was obtained by calculating the fundamental matrix from the projection matrices~\cite{hartley2003multiple}. Correspondences were considered as inliers if the symmetric epipolar distance was smaller than $1.0$ pixel. All image pairs  with less than $20$ inliers found were discarded. In total, $467$ image pairs were used in the evaluation. 

The results are reported in the $6$th block of Table~\ref{tab:dataset_comparison}. The reason of the high processing time is that the mean inlier ratio is relatively low ($27\%$) and there are many correspondences, $2323$, on average.
GC-RANSAC obtains the most accurate essential matrices both in the wall-clock time limited and solution confidence above $95\%$ experiments. A significant drop can be seen in accuracy for all methods if a time limit is given.

\paragraph{Estimation of Affine Transformation.}  

The {\fontfamily{cmtt}\selectfont SZTAKI Earth Observation} dataset\footnote{\url{http://mplab.sztaki.hu/remotesensing}} \cite{benedek2009change} ($83$ image pairs of size $320 \times 240$) was used to test estimation of affine transformations. The dataset contains images of busy road scenes taken from a balloon. Due to the altitude of the balloon, the image pair relation is well approximate by an affine transformation. Point correspondences were detected by the SIFT detector. For ground truth, $20$ inliers were selected manually. Point pairs with the distance from the ground truth affine transformation lower than $1.0$ pixel were defined as inliers.

The estimation results are shown in the $7$th block of Table~\ref{tab:dataset_comparison}. The reported geometric error is $| \textbf{A} \textbf{p}_1 - \textbf{p}_2 |$, where $\textbf{A}$ is the estimated affine transformation and $\textbf{p}_k$ is the point in the $k$th image ($k \in \{1,2\}$).  It can be seen that the methods obtained fairly similar results, however, GC-RANSAC is slightly more accurate. It is marginally slower due to the neighborhood computation. However, it is still faster than real time.  

\begin{table*}
\center
\caption{ Fundamental matrix estimation applied to {\fontfamily{cmtt}\selectfont kusvod2} (24 pairs), {\fontfamily{cmtt}\selectfont AdelaideRMF} (19 pairs) and {\fontfamily{cmtt}\selectfont Multi-H} (4 pairs) datasets, homography estimation on {\fontfamily{cmtt}\selectfont homogr} (16 pairs) and {\fontfamily{cmtt}\selectfont EVD} (15 pairs) datasets, essential matrix estimation on the {\fontfamily{cmtt}\selectfont strecha} dataset (467 pairs), and affine transformation estimation on  the {\fontfamily{cmtt}\selectfont SZTAKI Earth Observation} benchmark (52 pairs). Thus the methods were tested on total on $597$ image pairs.
The datasets, the problem ($\mathbf{F}/\mathbf{H}/\textbf{E}/\textbf{A}$), the number of the image pairs ($\#$) and the reported properties are shown in the first three columns. The next five report the results at $99\%$ confidence with a time limit set to $60$ FPS, i.e.\ the run is interrupted after $1 / 60$ secs (EP-RANSAC is removed since it cannot be applied in real time). For the other columns, there was no time limit but the confidence was set to $95\%$. Values are the means of $1000$ runs. LO is the number of local optimizations and the number of graph-cut runs are shown in brackets. The geometric error ($\mathcal{E}$, in pixels) of the estimated model w.r.t.\ the manually selected inliers is written in each second row; the mean processing time ($\mathcal{T}$, in milliseconds) and the required number of samples ($\mathcal{S}$) are written in every $3$th and $4$th rows. The geometric error is the Sampson distance for $\mathbf{F}$ and $\mathbf{E}$, and the projection error for $\mathbf{H}$ and $\mathbf{A}$. }
  	\resizebox{0.99\linewidth}{!}{\begin{tabular}{| l | l | r || r | r | r | r | r || r | r | r | r | r | r |  }
    \hline
 	 	 \multicolumn{3}{|c||}{} & \multicolumn{5}{c||}{Approx. 60 FPS (or 99\% confidence)} & \multicolumn{6}{c|}{Confidence 95\%} \\ 
    \hline 
 	 	 \multicolumn{3}{|c||}{} & \multicolumn{1}{c|}{RSC} & \multicolumn{1}{c|}{LO} & \multicolumn{1}{c|}{LO$^+$} & \multicolumn{1}{c|}{LO'} & \multicolumn{1}{c||}{\textbf{GC}} & \multicolumn{1}{c|}{RSC} & \multicolumn{1}{c|}{LO} & \multicolumn{1}{c|}{LO$^+$} & \multicolumn{1}{c|}{LO'} & \multicolumn{1}{c|}{EP-RSC} & \multicolumn{1}{c|}{\textbf{GC}} \\ 
    \hline    
 	 	\multirow{4}{*}{\rot{{\fontfamily{cmtt}\selectfont \footnotesize kusvod2}}} & \multirow{4}{*}{\rot{$\mathbf{F}$, $\#24$}} & LO & \emptycol & 2 & 2 & 2 & 1 (3) & \emptycol & 1 & 1 & 1 & \emptycol & 2 (3) \\
 	 	 & & $\mathcal{E}$ & 5.01 & 4.95 & 4.97 & 5.02 & \textbf{4.65} & 5.18 & 5.08 & 5.03 & 5.22 & 7.87 & \textbf{4.69} \\
 	 	 & & $\mathcal{T}$ & \phantom{xx}6.2 & \phantom{x}6.1 & \phantom{x}6.3 & \phantom{x}5.9 & \phantom{x}\textbf{4.6} & 4.9 & \phantom{x}5.2 & \phantom{x}5.1 & \phantom{x}4.9 & 439.9 & \phantom{x}\textbf{3.6} \\
 	 	 & & $\mathcal{S}$ & 117 & 96 & 99 & 111 & \textbf{70} & 93 & 76 & 78 & 87 & \emptycol & \textbf{53} \\
    \hline 
 	 	\multirow{4}{*}{\rot{{\fontfamily{cmtt}\selectfont \footnotesize Adelaide}}} & \multirow{4}{*}{\rot{$\mathbf{F}$, $\#19$}} & LO & \emptycol & 2 & 2 & 2 & 1 (3) & \emptycol & 2 & 2 & 3 & \emptycol & 2 (4) \\
 	 	 & & $\mathcal{E}$ & 0.55 & 0.53 & 0.52 & 0.55 & \textbf{0.50} & 0.44 & 0.45 & \textbf{0.43} & 0.44 & 0.71 & \textbf{0.43} \\
 	 	 & & $\mathcal{T}$ & 14.2 & 14.8 & 14.9 & \textbf{14.1} & 18.9 & 262.7 & \textbf{194.2} & 210.9 & 237.1 & 2\;121.9 & 227.1 \\
 	 	 & & $\mathcal{S}$ & 124 & \textbf{113} & \textbf{113} & 122 & 116 & 1\;363 & 1\;126 & 1\;205 & 1\;305.00 & \emptycol & \textbf{1\;115} \\
    \hline
 	 	\multirow{4}{*}{\rot{{\fontfamily{cmtt}\selectfont \footnotesize Multi-H}}} & \multirow{4}{*}{\rot{$\mathbf{F}$, $\#4$}} & LO & \emptycol & 1 & 1 & 1 & 1 (3) & \emptycol & 2 & 1 & 2 & \emptycol & 1 (3) \\
 	 	&  & $\mathcal{E}$ & 0.35 & 0.34 & 0.34 & 0.34 & \textbf{0.32} & 0.33 & 0.33 & 0.33 & 0.34 & 0.44 & \textbf{0.32} \\
 	 	 & & $\mathcal{T}$ & \textbf{10.3} & 11.5 & 11.1 & \textbf{10.3} & 14.6 & 12.8 & 15.1 & 14.1 & \textbf{12.4} & 2\;371.8 & 36.0 \\
 	 	 & & $\mathcal{S}$ & 83 & 76 & 76 & 82 & \textbf{74} & 107 & 89 & 90 & 100 & \emptycol & \textbf{78} \\
    \hline        
 	 	\multirow{4}{*}{\rot{{\fontfamily{cmtt}\selectfont \footnotesize EVD}}} & \multirow{4}{*}{\rot{$\mathbf{H}$, $\#15$}} & LO & \emptycol & 2 & 2 & 2 & 2 (2) & \emptycol & 4 & 4 & 4 & \emptycol & 3 (6) \\
 	 	 & & $\mathcal{E}$ & 1.53 & 1.63 & \textbf{1.51} & 1.58 & 1.53 & 0.96 & 0.95 & 0.95 & 0.96 & 1.17 & \textbf{0.92} \\
 	 	 & & $\mathcal{T}$ & 16.8 & 18.3 & 18.0 & \textbf{16.8} & 19.2 & 247.3 & 248.0 & 251.3 & \textbf{247.0} & $>10^4$ & 249.9 \\
 	 	 & & $\mathcal{S}$ & 320 & \textbf{298} & 301 & 318 & 301 & 4\;303 & \textbf{4\;203} & 4\;248 & 4\;291 & \emptycol & 4\;204 \\         
    \hline
 	 	\multirow{4}{*}{\rot{{\fontfamily{cmtt}\selectfont \footnotesize homogr}}} & \multirow{4}{*}{\rot{$\mathbf{H}$, $\#16$}} & LO & \emptycol & 2 & 2 & 2 & 1 (3) & \emptycol & 2 & 2 & 2 & \emptycol & 1 (4) \\
 	 	 & & $\mathcal{E}$ & 0.53 & 0.53 & 0.53 & 0.53 & \textbf{0.51} & 0.50 & 0.50 & 0.49 & 0.50 & 0.58 & \textbf{0.47} \\
 	 	 & & $\mathcal{T}$ & 7.1 & 10.4 & 9.8 & \textbf{7.1} & 7.6 & 17.1 & 10.1 & 9.9 & 8.5 & 3\;339.7
 & \textbf{7.9} \\
 	 	 & & $\mathcal{S}$ & 193 & 175 & 175 & 189 & \textbf{159} & 450 & 212 & 214 & 226 & \emptycol & \textbf{165} \\
    \hline
 	 	\multirow{4}{*}{\rot{{\fontfamily{cmtt}\selectfont \footnotesize strecha}}}  & \multirow{4}{*}{\rot{$\mathbf{E}$, $\#467$}} & LO & \emptycol & 1 & 1 & 1 & 1 (1) & \emptycol & 7 & 7 & 7 & \emptycol & 7 (7) \\
 	 	 & & $\mathcal{E}$ & 11.81 & 12.34 & 12.07 & 12.12 & \textbf{11.6} & 3.03 & 2.95 & 2.94 & 2.87 & 3.32 & \textbf{2.83} \\
 	 	 & & $\mathcal{T}$ & 11.6 & 17.3 & \textbf{17.2} & \textbf{17.2} & 17.3 & 3\;581.9 & 3\;638.5 & 3\;648.4 & 3\;570.0 & $>10^6$ & \textbf{3\;466.4} \\
 	 	 & & $\mathcal{S}$ & 31 & \textbf{30} & 31 & 31 & \textbf{30} & 3\;654 & 3\;646 & \textbf{3\;634} & 3\;653 & \emptycol & 3\;651 \\
    \hline
 	 	\multirow{4}{*}{\rot{{\fontfamily{cmtt}\selectfont \footnotesize SZTAKI}}}  & \multirow{4}{*}{\rot{$\mathbf{A}$, $\#52$}} & LO & \emptycol & 1 & 1 & 1 & 1 (3) & \emptycol & 1 & 1 & 1 & \emptycol & 1 (3) \\
 	 	 & & $\mathcal{E}$ & 0.41 & 0.41 & 0.41 & 0.41 & \textbf{0.40} & 0.45 & 0.46 & 0.44 & 0.45 & 0.48 & \textbf{0.41} \\
 	 	 & & $\mathcal{T}$ & 3.5 & \textbf{3.2} & \textbf{3.2} & \textbf{3.2} & 10.3 & \textbf{1.7} & \textbf{1.7} & \textbf{1.7} & \textbf{1.7} & 4\;718.2 & 10.2 \\
 	 	 & & $\mathcal{S}$ & 26 & 26 & 26 & 26 & 26 & \textbf{9} & \textbf{9} & \textbf{9} & \textbf{9} & \emptycol & \textbf{9} \\
    \hline
\end{tabular}}
\label{tab:dataset_comparison}
\end{table*}

\paragraph{Convergence from a Not-All-Inlier Sample.}  
Table~\ref{tab:not_all_inl_comparison} reports the frequencies when a ``not-all-inlier'' sample led to the correct model. For lines ($\mathbf{L}$), it is computed using $1000$ runs on each outlier (100, 500 and 1000) and noise level (from $0.0$ up to $9.0$ pixels). Thus $15000$ runs were performed. A minimal sample is counted as a ``not-all-inlier'' if it contains at least one point farther from the ground truth model than the ground truth noise $\sigma$.

For fundamental matrices ($\mathbf{F}$), the frequencies of success from a ``not-all-inlier'' sample are computed as the mean of $1000$ runs on all pairs of the {\fontfamily{cmtt}\selectfont AdelaideRMF} dataset. In this dataset, all inliers are labeled manually, thus it is easy to check whether a sample point is inlier or not. 

\begin{figure}
  \centering
  \begin{subfigure}[m]{1.00\columnwidth}
  	\centering
    \raisebox{0.205\columnwidth}{(a)} \includegraphics[width=0.85\columnwidth]{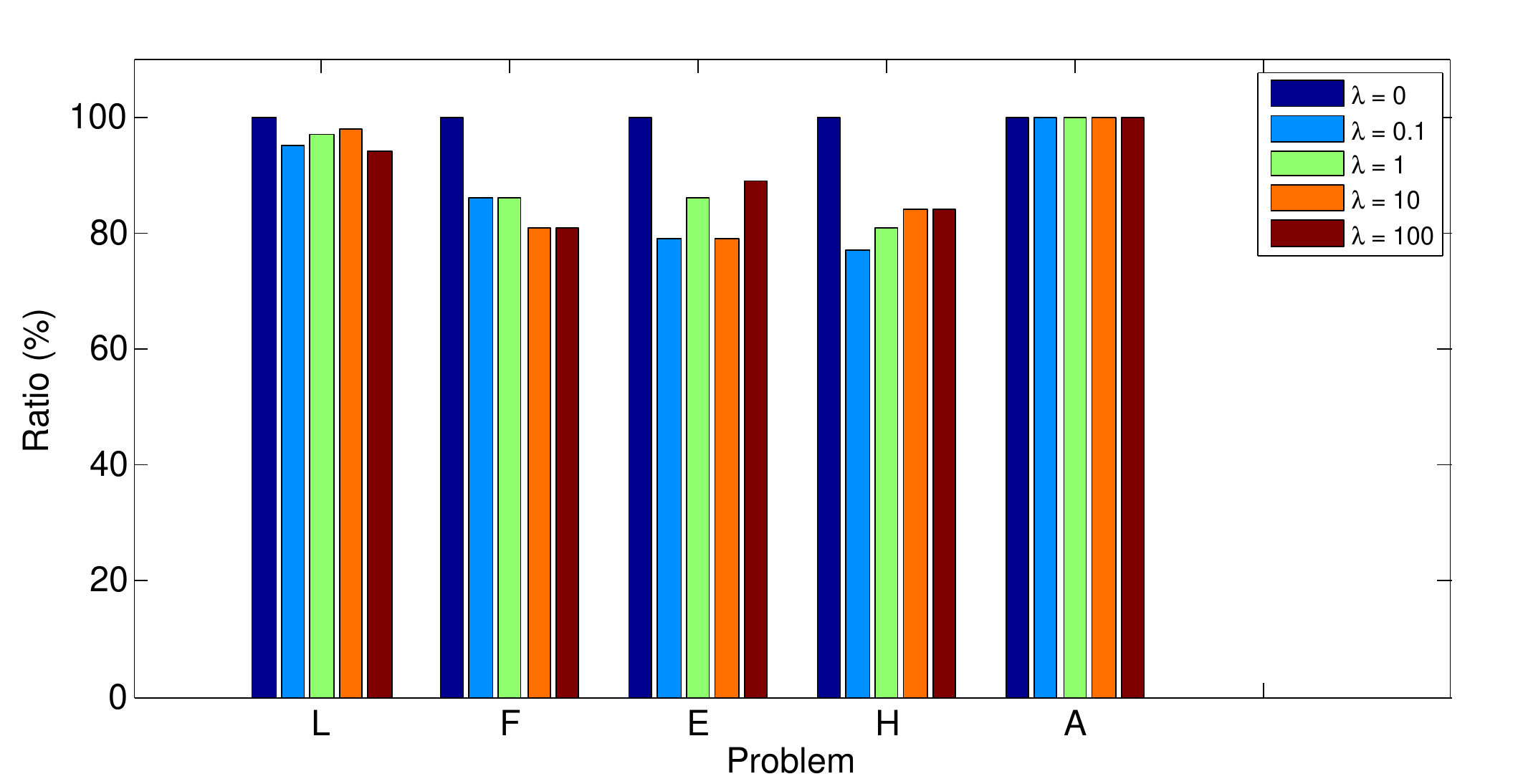}
  \end{subfigure}
  \begin{subfigure}[b]{1.00\columnwidth}
  	\centering
  	\raisebox{0.205\columnwidth}{(b)}  \includegraphics[width=0.85\columnwidth]{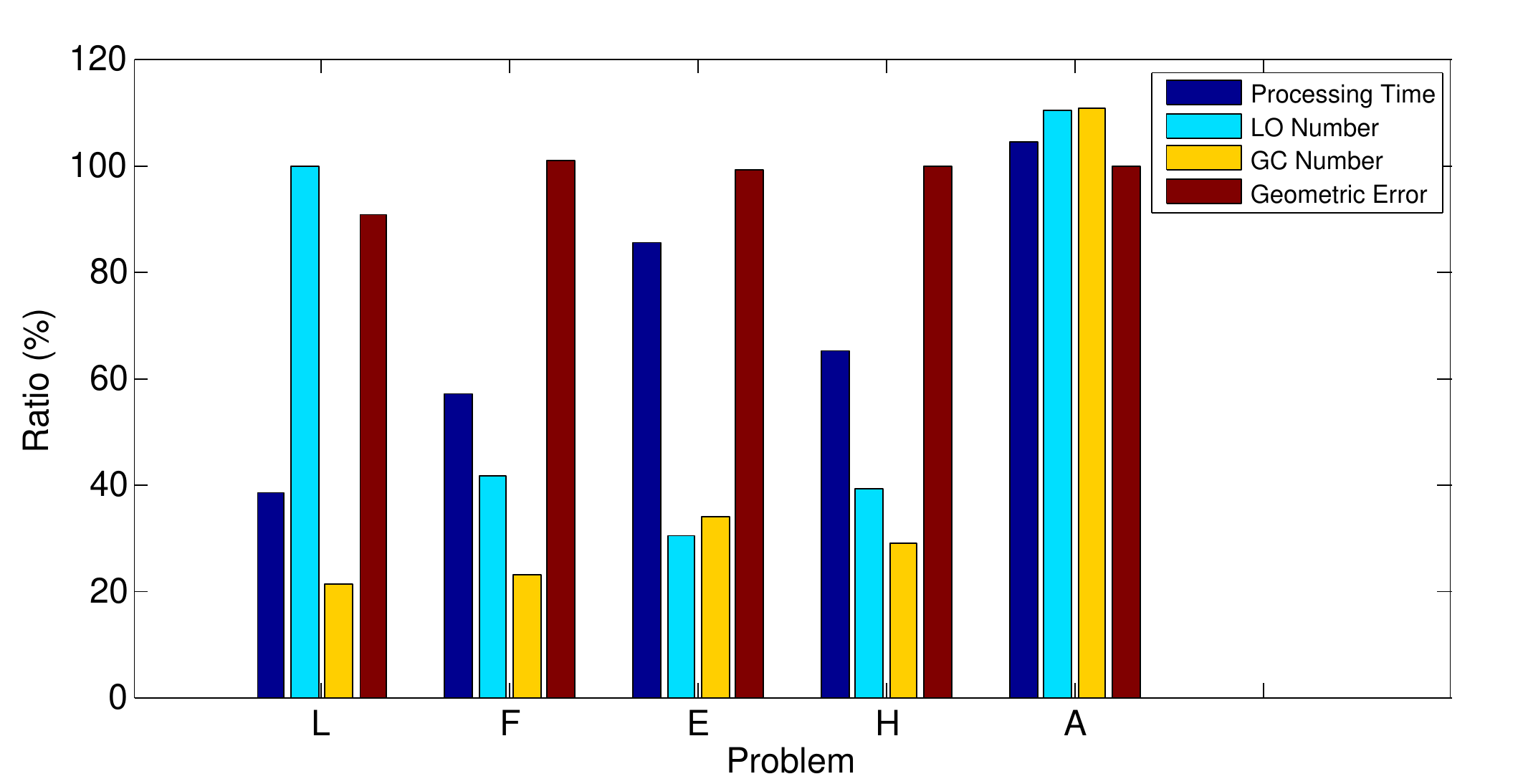}
  \end{subfigure}
  \begin{subfigure}[b]{1.00\columnwidth}
  	\centering
  	\raisebox{0.205\columnwidth}{(c)}  \includegraphics[width=0.85\columnwidth]{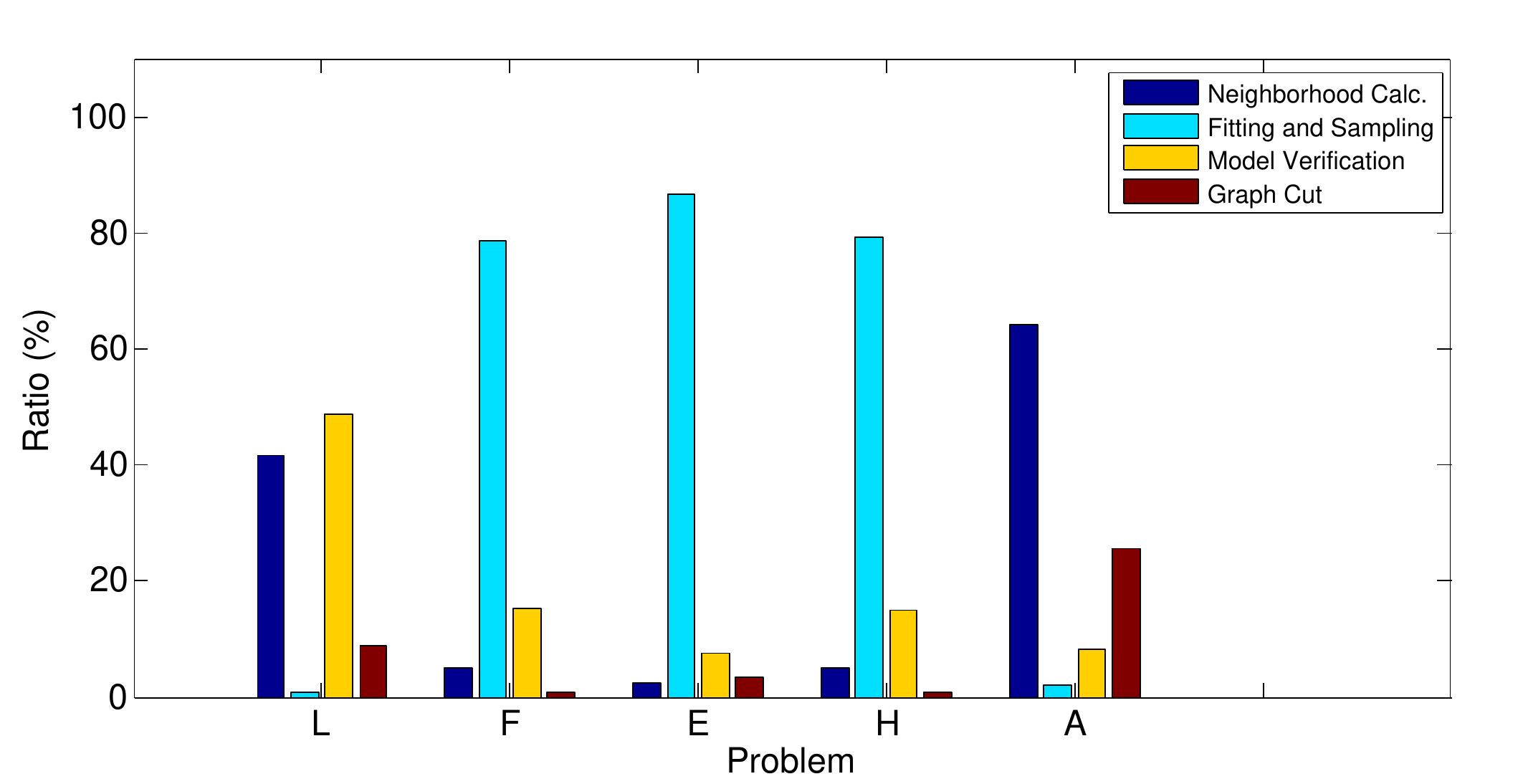}
  \end{subfigure}
  \caption{ (\textbf{a}) The effect of the $\lambda$ choice weighting the spatial term. The ratio of the geometric error (in percentage) compared to the $\lambda = 0$ case (no spatial coherence) for each problem ($\textbf{L}$ -- lines, $\textbf{F}$ -- fundamental matrix, $\textbf{E}$ -- essential matrix, $\textbf{H}$ -- homography, $\textbf{A}$ -- affine transformation). (\textbf{b}) The effect of replacing the iteration limit before the first LO applied with the proposed criterion, i.e.\ the confidence radically increases. The ratios (in percentage) of each property of the proposed and that of standard approaches. (\textbf{c}) The breakdown of the processing times in percentage w.r.t.\ the total runtime. All values were computed as the mean of all tests. \textit{Best viewed in color.}} 
  \label{fig:processing_bars}
\end{figure}

\paragraph{Evaluation of the $\lambda$ setting.} 

To evaluate the effect of the $\lambda$ parameter balancing the spatial coherence term, we applied GC-RANSAC to all problems with varying $\lambda$. The evaluated values are: (i) $\lambda = 0$, which turns off the spatial coherence term, (ii) $\lambda = 0.1$, (iii) $\lambda = 1$, (iv) $\lambda = 10$, and (v) $\lambda = 100$. Fig.~\ref{fig:processing_bars}{\color{red} a} shows the ratio of the geometric errors for   $\lambda \neq 0 $ and $\lambda = 0$ (in percent). For all investigated non-zero $\lambda$ values, the error is lower than for $\lambda = 0$. Since $\lambda = 0.1$ led to the most accurate results on average, we chose this setting in the tests. 

\paragraph{Evaluation of the criterion for the local optimization.}

The proposed criterion (Eq.~\ref{eq:criterion}) ensuring that local optimization is applied only to the most promising model candidates is tested in this section. We applied GC-RANSAC to all problems combined with the proposed and the standard approaches. The standard technique sets an iteration limit (default value: $50$) and the LO procedure is afterwards applied to all models that are so far the best. Fig.~\ref{fig:processing_bars}{\color{red} b} reports the ratio of each property (processing time -- dark blue, LO -- light blue,  and GC steps -- yellow, geometric error -- brown) of the proposed and standard approaches. The new criterion leads to significant improvement in the processing time with no deterioration in accuracy. 

\paragraph{Processing Time.}  
Fig.~\ref{fig:processing_bars}{\color{red} c} shows the breakdown of the processing times of GC-RANSAC applied to each problem. The time demand of the neighborhood computation (dark blue) linearly depends on the point number. The light blue one is the time demand of the sampling and model fitting step, the yellow and brown bars show the model verification (support computation) and the proposed local optimization step, respectively. The sampling and model fitting part dominates the process.

\section{Conclusion}

GC-RANSAC was presented. It is more geometrically accurate than state-of-the-art methods. It runs in real-time for many problems at a speed approximately equal to the less accurate alternatives. It is much simpler to implement in a reproducible manner than any of the competitors (RANSAC's with local optimization). Its local optimization step is globally optimal for the so-far-the-best model parameters. We also proposed a criterion for the application of the local optimization step. This criterion leads to a significant improvement in processing time with no deterioration in accuracy. GC-RANSAC can be easily inserted into USAC~\cite{raguram2013usac} and be combined with its ''bells and whistles`` like PROSAC sampling, degeneracy testing and fast evaluation with early termination.

{\small
\bibliographystyle{ieee}
\bibliography{egbib}
}

\end{document}